\documentclass[runningheads]{llncs}
\usepackage[T1]{fontenc}
\usepackage{ifpdf}
\ifpdf
  \usepackage{graphicx}
\else
  \usepackage[dvipdfmx]{graphicx}
\fi
\usepackage{caption}
\usepackage{array}
\usepackage{booktabs}
\usepackage{amsmath}
\usepackage{amssymb,bm}
\usepackage{multirow, multicol}
\usepackage{xurl}
\ifpdf
  \usepackage[table]{xcolor}
\else
  \usepackage[dvipdfmx,table]{xcolor}
\fi
\usepackage{colortbl}
\definecolor{testcolor}{rgb}{1, 1, 0.6}
\definecolor{myred}{HTML}{B85450}
\definecolor{myblue}{HTML}{6C8EBF}
\definecolor{flickr8k-cf}{RGB}{129,190,142}
\definecolor{flickr8k-ex}{RGB}{211,123,126}
\definecolor{TitleColor}{gray}{0.95}
\definecolor{LightCyan}{rgb}{0.88,0.95,1}
\definecolor{LightPink}{HTML}{FAE6E7}

\makeatletter
\renewcommand\paragraph{\@startsection{paragraph}{4}{\z@}%
  {0.5em}%
  {-1em}%
  {\normalfont\bfseries}}
\makeatother

\begin{document}
\title{ABMAMBA: Multimodal Large Language Model with Aligned Hierarchical Bidirectional Scan for Efficient Video Captioning}
\titlerunning{ABMAMBA}
%
\author{Daichi Yashima\inst{1,3}\orcidID{0009-0005-2087-2038} \and
Shuhei Kurita \inst{2,3}\orcidID{0000-0001-7415-3120} \and
Yusuke Oda\inst{3}\orcidID{0009-0009-1681-7123} \and
Shuntaro Suzuki\inst{1}\orcidID{0009-0008-5564-3835}\and
Seitaro Otsuki\inst{1}\orcidID{0009-0009-8071-6060}\and
Komei Sugiura\inst{1}\orcidID{0000-0002-0261-0510}
}
\authorrunning{D. Yashima et al.}
\institute{
Keio University,  
3-14-1, Kohoku Ward, Yokohama, Kanagawa 223-8522, Japan  
\and
National Institute of Informatics,  
2-1-2 Hitotsubashi, Chiyoda,
Tokyo 101-8430, Japan  
\and
National Institute of Informatics Research and Development Center for Large Language Models,
1-1-1, Hitotsubashi, Chiyoda, Tokyo 100-0003, Japan   \\
 }
\maketitle              
\begin{abstract}
In this study, we focus on video captioning by fully open multimodal large language models (MLLMs).
The comprehension of visual sequences is challenging because of their intricate temporal dependencies and substantial sequence length.
The core attention mechanisms of existing Transformer-based approaches scale quadratically with the sequence length, making them computationally prohibitive.
To address these limitations, we propose Aligned Hierarchical Bidirectional Scan Mamba (ABMamba), a fully open MLLM with linear computational complexity that enables the scalable processing of video sequences.
ABMamba extends Deep State Space Models as its language backbone, replacing the costly quadratic attention mechanisms, and employs a novel Aligned Hierarchical Bidirectional Scan module that processes videos across multiple temporal resolutions.
On standard video captioning benchmarks such as VATEX and MSR-VTT, ABMamba demonstrates competitive performance compared to typical MLLMs while achieving approximately three times higher throughput. Our project page is available at \url{https://v8utn.kinsta.page}.

\keywords{Video MLLM  \and Deep SSM \and Video captioning.}
\end{abstract}

\section{Introduction}
Recent advances in multimodal large language models (MLLMs)~\cite{reid2024gemini,llavaonevision} have enhanced multimodal understanding capabilities beyond static image perception to the comprehension of video content with complex spatiotemporal dynamics.
These capabilities have led to the use of MLLMs for various downstream tasks, including video QA, video captioning, and text--video retrieval~\cite{tang2023video}.
\begin{figure}[t]
    \centering
    \includegraphics[width=0.7\linewidth]{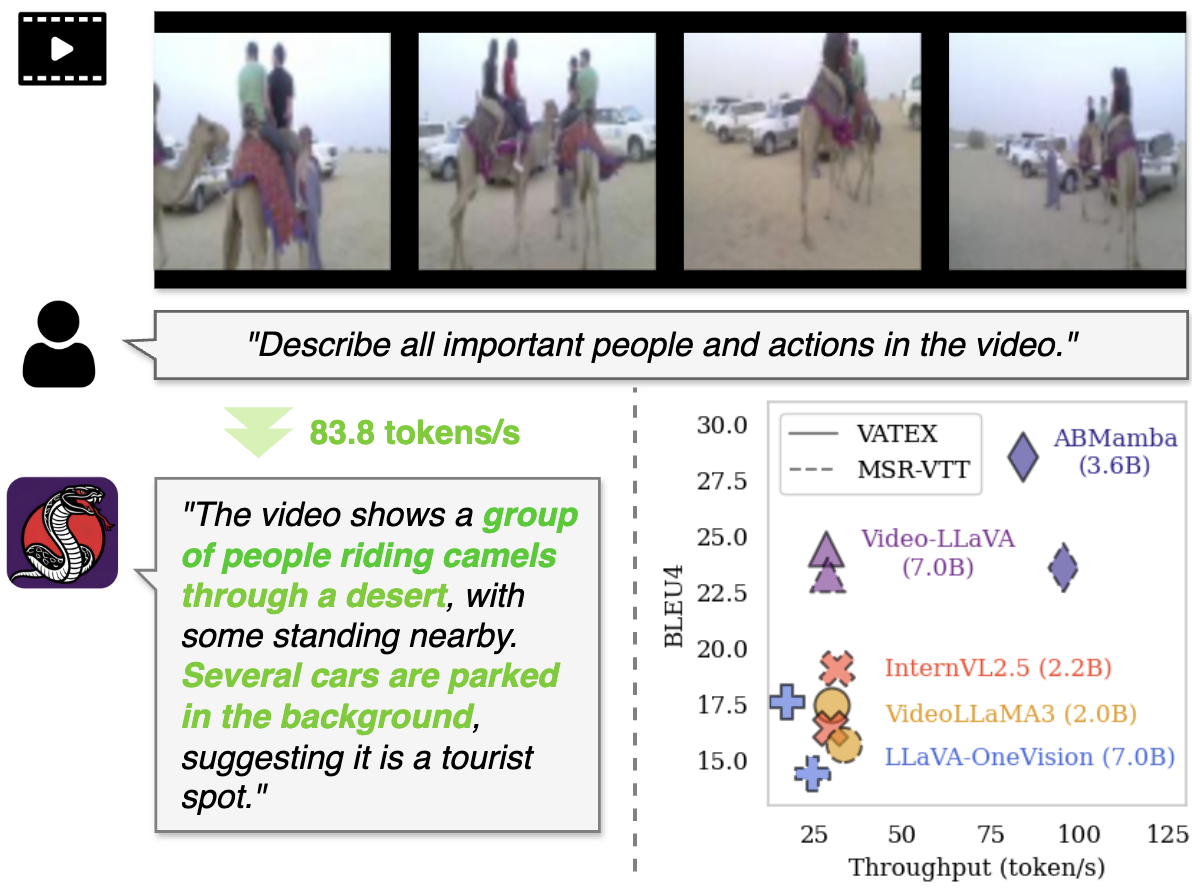}
    \caption{A typical use case of ABMamba. ABMamba generates relevant and descriptive captions for a given video accompanied by a natural language prompt. {Green} text highlights the correctly captured contextual information.}
    \label{fig:eye-catch}
    \vspace{-4mm}
\end{figure}
Video captioning has practical utility across a wide range of domains, such as video summarization, visual scene interpretation in domestic service robots, and vision accessibility applications~\cite{tang2023video,nguyen-etal-2024-video}.
Current MLLMs have a limited video captioning ability when processing long-duration inputs or operating in real time. Addressing these limitations would enable broader applications across several domains. 

In this study, we focus on video captioning tasks by fully open MLLMs.
Here `fully open' refers to the availability of open datasets, code, and trained model weights.
Fig.~\ref{fig:eye-catch} illustrates a typical use case of ABMamba.
The input consists of a video and a natural language prompt.
Given this input, the model generates relevant and descriptive captions for the video.

This task is challenging because the visual input  has intricate temporal dependencies and a substantial sequence length.
Consequently, the naive application of Transformer-based approaches often results in prohibitive computational costs, as their core attention mechanisms scale quadratically with respect to the sequence length.
In response, prior approaches often resort to input compression (e.g., downsampling or learned projections), which inevitably sacrifices fine-grained temporal details.

To address these issues, we propose ABMamba, a novel, fully open MLLM based on Deep State Space Models (Deep SSMs).
By replacing quadratic attention mechanisms of the conventional language backbone with Deep SSMs, ABMamba achieves efficient temporal modeling with linear computational complexity, allowing the scalable processing of long video sequences.
To capture the intricate temporal dynamics, we employ a novel Aligned Hierarchical Bidirectional Scan (AHBS) module that 
propagates information both forward and backward across multiple resolutions, overcoming the coarse summarization and loss of sequential cues associated with the simple downsampling or projection strategies.
Moreover, we make ABMamba fully open to facilitate community-wide efforts toward fundamental improvements in core algorithms.

Our main contributions are as follows:
\begin{itemize}
    \item We propose ABMamba, a fully open Deep SSM-based MLLM that achieves efficient temporal modeling of videos with sub-quadratic computational scaling in sequence length.
    \item We introduce a novel AHBS module that processes videos temporally across multiple resolutions. 
    This module enables the complex and intricate temporal dynamics inherent in videos to be effectively captured.
    \item While achieving competitive performance across most evaluation metrics, ABMamba delivers 2–3 times the inference speed of baseline methods.
\end{itemize}

\section{Related Work}
Rapid advances in MLLMs have been extensively reviewed, with recent surveys categorizing major trends in model architectures and training paradigms~\cite{liang2024survey,li2024multimodal}.
While Transformer-based architectures remain the dominant backbones of current MLLMs, recent studies have shown that Deep SSMs outperform Transformers in sequence modeling tasks; a diverse array of variants has been systematically reviewed~\cite{patro2024mamba,wang2024state}.
In parallel, extensive efforts have been made to adapt Deep SSMs for visions tasks~\cite{Liu2024VisionMA,xu2024visual,rahman2024mamba}.

\paragraph{Multimodal large language models.}
Although many early breakthroughs centered on handling image- and text-based tasks, a growing body of work now targets video understanding~\cite{tang2023video}.
Nevertheless, the modeling of temporal dependencies in video and scaling to long sequences remain open challenges.
Recent video MLLMs (e.g.,~\cite{videollava,llavaonevision,molmo,video-XL}) address this by incorporating frame-wise vision encoders with a lightweight projection and instruction-tuned datasets (e.g.,~\cite{videollava,llavaonevision,llavavideodata}).  
For instance, LLaVA-OneVision~\cite{llavaonevision} and Molmo~\cite{molmo} project each frame independently using vision encoders (e.g., ~\cite{clip,zhai23iccv}) followed by simple MLPs, but lack mechanisms for integrating temporal dependencies.
Thus, it is difficult to capture causal relations across frames.
In contrast, models such as Video-XL~\cite{video-XL}  introduce a chunk-wise summarization strategy using latent tokens.
This enables efficient long-video processing but sacrifices per-frame granularity and fine temporal detail.

Conventional Transformer-based MLLMs encounter scalability limitations when faced with long sequences because of their quadratic computational complexity with respect to the sequence length. 
Existing vision--language projection methods, ranging from complex Q-Formers~\cite{videollava} to efficient yet potentially limited MLPs and pooling~\cite{maaz24acl}, exhibit a trade-off between architectural complexity and representational power for cross-modal alignment. 
To address these challenges, our model extends Deep SSMs with linear complexity, inherently improving the efficiency for long sequences. 
The proposed AHBS module resolves this tension by performing multi-resolution, bidirectional token scanning that preserves fine-grained temporal information while retaining sub-quadratic computational cost during visual integration. 

Proprietary MLLMs~\cite{achiam2023gpt,reid2024gemini} have demonstrated impressive video comprehension performance.
However, the growing demand for transparency, reproducibility, and accessibility highlights the importance of developing open models.
These demands have driven the development of open MLLMs, such as the LLaVA series~\cite{llava,llavavideodata,llavaonevision} and Pangea~\cite{yue2024pangea}, offering transparent alternatives for both academic research and real-world deployment.
Our work contributes to this growing ecosystem by introducing a fully open MLLM based on Deep SSMs.

\paragraph{Deep state space models.}
Although Transformer-based architectures are currently the dominant method of sequence modeling, their quadratic complexity has motivated extensive research into more efficient alternatives that maintain comparable performance~\cite{sun2023retentivenetworksuccessortransformer,katharopoulos-et-al-2020}.
Among these, Deep SSMs~\cite{gu22iclr,s5smith2023simplified,gu2024mamba,mamba2} have gained increasing attention for their strong capabilities in long-range sequence modeling~\cite{patro2024mamba,wang2024state}.
A key advantage of SSMs lies in their dual formulation: they can be implemented as recurrent neural networks to support efficient autoregressive inference, while also being reformulated for parallel sequence processing during training~\cite{gu2021combining,gu22iclr}.

A recent selective SSM, Mamba~\cite{gu2024mamba} has demonstrated superior performance to Transformers in certain language modeling tasks.
Building on this foundation, LLMs with hybrid architectures combining Mamba and Transformer have been proposed, including Jamba~\cite{lenz2025jamba} and Nemotron-H~\cite{nvidia2025nemotronhfamilyaccurateefficient}.
Furthermore, recent models such as Cobra~\cite{zhao2025cobra}, VL-Mamba~\cite{qiao2024vlmamba}, and EMMA~\cite{xing2025emma} have adopted Deep SSMs as the core backbone architecture for the MLLM, achieving competitive accuracy with fewer parameters and faster inference.  
To date, however, these models have been restricted to static images and have not fully addressed temporal modeling.

\paragraph{Deep SSMs for vision.}

Studies that apply Deep SSMs to visual tasks have reported notable success~\cite{Liu2024VisionMA,xu2024visual,rahman2024mamba}.
In particular, S4ND~\cite{s4nd} employs S4~\cite{gu22iclr} to scan images vertically and horizontally, using the outer product of the resulting vectors as convolutional kernels.
Similarly, 2D-SSM~\cite{baron2024a} introduces state matrices that explicitly distinguish state transitions along the vertical and horizontal axes.
However,~\cite{visionmamba} identified two major challenges in the application of Deep SSMs to vision: unidirectional modeling and the lack of location awareness.
To address these issues, Vim~\cite{visionmamba} employs a bidirectional scanning method based on Mamba~\cite{gu2024mamba}.
Various scanning strategies, including the zigzag scan~\cite{hu2024zigma}, Hilbert scan~\cite{haoyang24neurips}, and cross scan~\cite{qiao2024vlmamba}, have been developed to better capture the non-causal structure inherent in visual data.

Furthermore, recent studies have attempted to apply Deep SSMs to long-sequence modeling for video processing~\cite{videomamba,patro2024simbasimplifiedmambabasedarchitecture,Zou_Guo_Hu_Ma_2025}.
For instance, VideoMamba~\cite{videomamba} uses a bidirectional scanning approach across patches from all frames within the video.
Nevertheless, these video-focused methods do not explicitly model the diverse temporal dynamics intrinsic to video data.
Moreover, as they are not integrated with language representations, they are not directly applicable to multimodal tasks such as video captioning.


%
\section{Method}

\subsection{Model Overview}
\label{sec:main-method}

We propose ABMamba, the first fully open video MLLM, based on Deep SSMs for capturing the intricate temporal dynamics inherent in videos.
Our method is inspired by recent advances in fully open MLLMs, which have shown remarkable capabilities in understanding and reasoning across different modalities. 
Compared with existing MLLMs, ABMamba differs in the following aspects~\cite{llava,llavaonevision,yue2024pangea,molmo}.
We specifically introduce an LLM backbone based on Mamba~\cite{gu2024mamba} and a novel AHBS module, which processes video temporally across multiple resolutions.
The AHBS module provides a modular approach for capturing temporal dependencies in video, a core challenge for many MLLMs when handling video inputs.
As a result, the AHBS module has the potential for broader adoption in other architectures.
Furthermore, our efficient video MLLM is broadly applicable to methods such as vision language action models~\cite{black2024pi_0,brohan2022rt,kim2024openvla} or fields such as autonomous driving~\cite{DriveGPT4,driveasyouspeak,tian2024drivevlm} because of its efficient processing of long-range temporal dependencies in video.


\begin{figure}[t]
    \centering
    \includegraphics[width=0.6\linewidth]{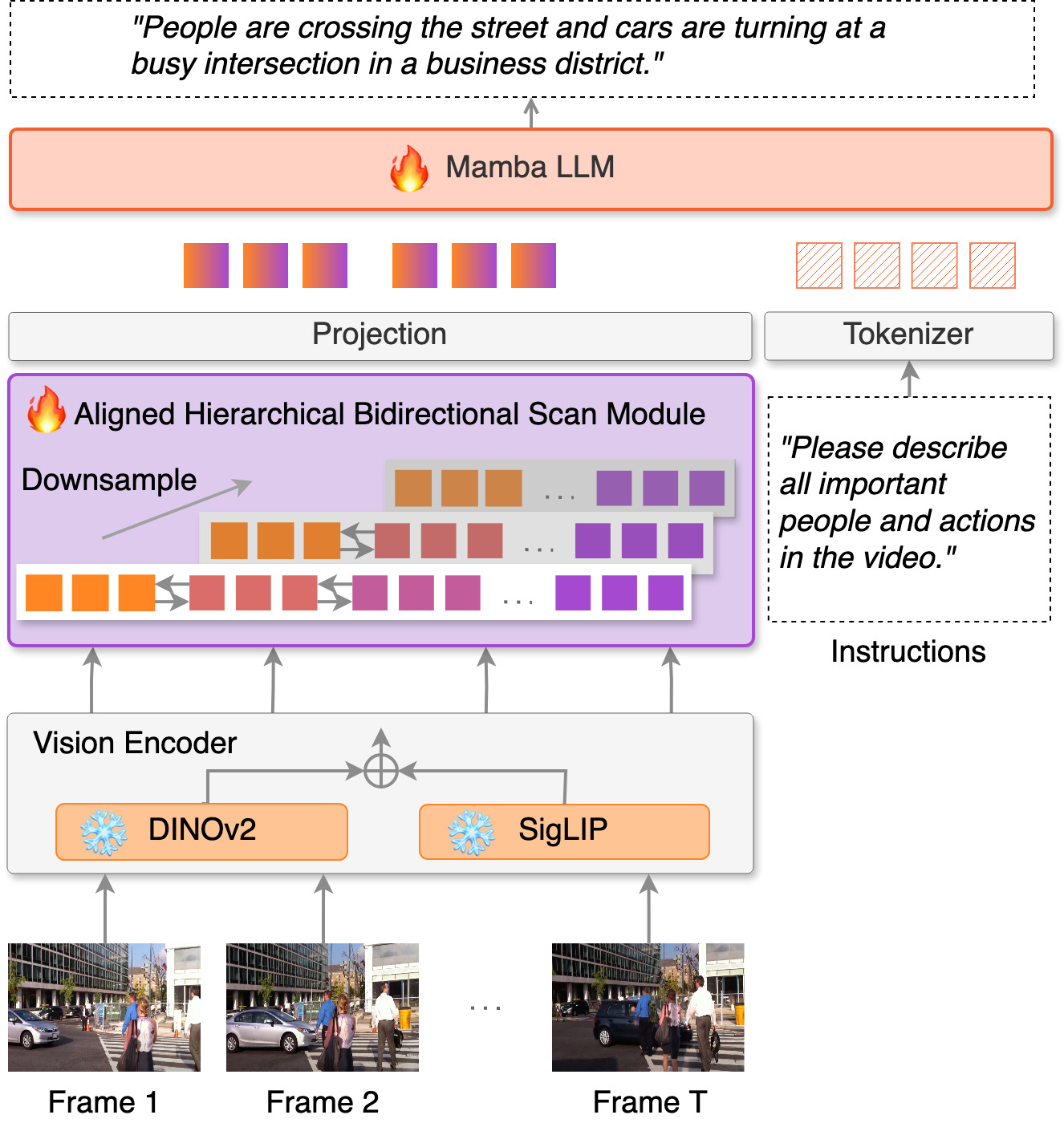}
    \caption{The architecture of ABMamba. Given a video with a language prompt, the model generates a caption that concisely describes the visual content. The model consists of a vision encoder, the aligned hierarchical bidirectional scan module, and a Mamba based LLM.}
    \label{fig:model}
 \vspace{-5mm}
\end{figure}
\begin{figure*}[t]
    \centering
    \includegraphics[width=\textwidth]{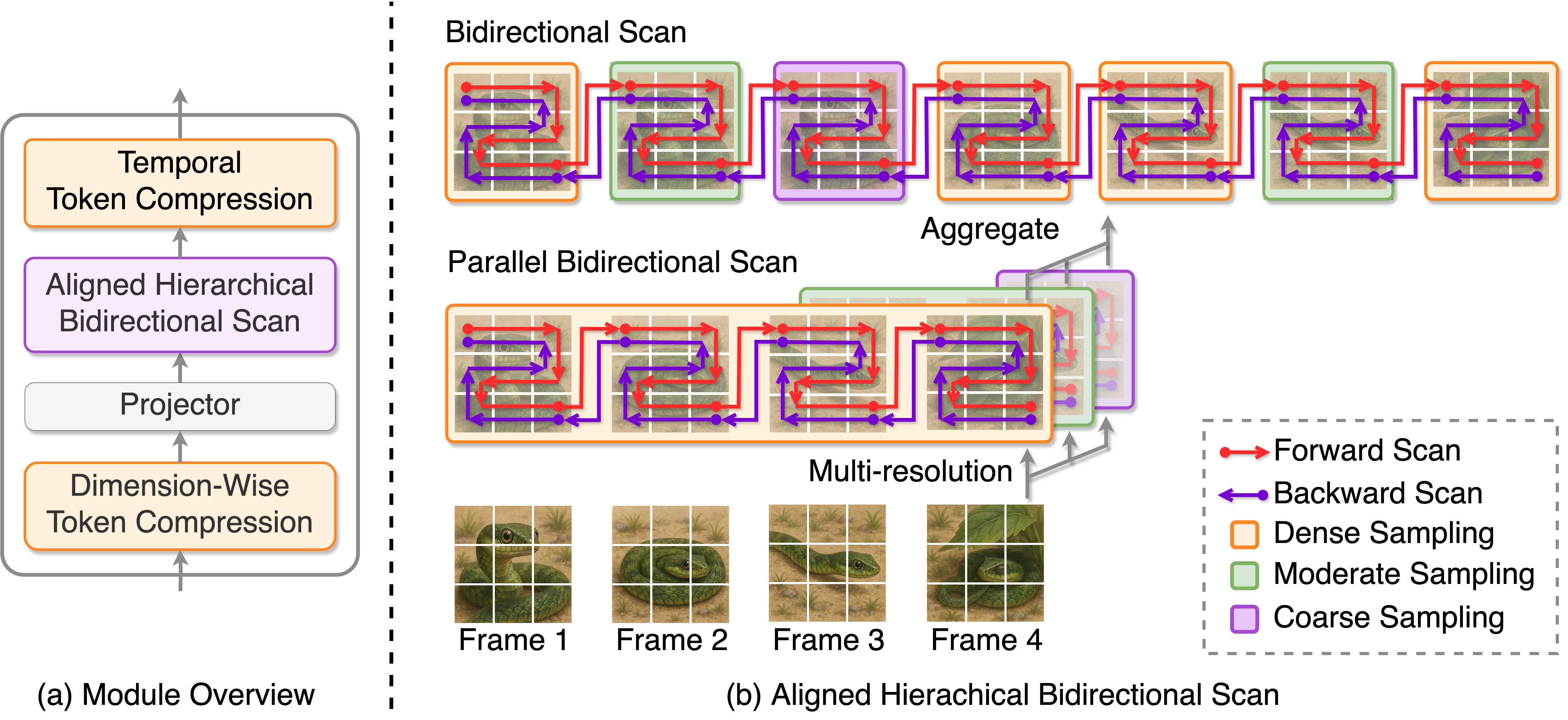}
    \caption{The overview of Aligned Hierarchical Bidirectional Scan (AHBS) module. (a) The module consists of a dimension-wise token compression, projector, AHBS, and temporal token compression. (b) The AHBS explicitly models intricate temporal dynamics through multi-resolution parallel bidirectional scan.}
    \label{fig:scan}
\vspace{-5mm}
\end{figure*}
Fig.~\ref{fig:model} shows the architecture of ABMamba.
Our model mainly consists of three modules: a vision encoder, the AHBS module, and an LLM specifically based on Mamba.

\subsection{Vision Encoder}
The inputs are defined as $\bm{x} = (\bm{x}_\text{vision}, \bm{x}_\text{txt})$, where
$\bm{x}_{\text{vision}} = (\bm{x}_{\text{vision}}^{(1)}, \bm{x}_{\text{vision}}^{(2)}, \dots,\allowbreak \bm{x}_{\text{vision}}^{(T)}), \bm{x}_{\text{vision}} \in \mathbb{R}^{T \times 3 \times H \times W}$
and $\bm{x}_\text{txt} \in \mathbb{R}^{V \times L}$ denote either a single image or video frame and a text input tokenized as one-hot vector, respectively.
Here, $T$, $H$, $W$, $V$, and $L$ denote the number of frames, frame height, frame width, vocabulary size, and sequence length, respectively.
A single image corresponds to the case where $T=1$.

We leverage the complementary strengths of SigLIP~\cite{zhai23iccv} and DINOv2~\cite{oquab24tmlr} as vision encoders. 
This design is motivated by recent advances in MLLMs which indicate that a dual vision encoder setup can significantly enhance visual understanding~\cite{zhao2025cobra,tong24neurips,goko2024task}. 
SigLIP has an image--text contrastive framework that provides robust semantic alignment, while the self-supervised approach of DINOv2 yields fine-grained visual features that capture subtle details.

Each frame $\bm{x}_\text{vision}^{(t)}$ is independently processed as follows: 
First, $\bm{x}_\text{vision}^{(t)}$ is divided into $N_\text{p}$ non-overlapping patches, each of size $p\times p$, where $N_\text{p} = (H\times W) / p^2$.
The patched images are then input to the two vision encoders to obtain $\bm{V}_{\text{SigLIP}}^{(t)} \in \mathbb{R}^{N_\text{p} \times d_\text{s}}$  and $\bm{V}_{\text{DINOv2}}^{(t)} \in \mathbb{R}^{N_\text{p} \times d_\text{d}}$ where $d_\text{s}$ and $d_\text{d}$ represent the output dimensions of the SigLIP and DINOv2 encoders, respectively.
Subsequently, $\bm{V}_{\text{SigLIP}}$ and $\bm{V}_{\text{DINOv2}}$ are concatenated along the feature dimension to form a unified representation: $\bm{V}^{(t)} = [\bm{V}_{\text{SigLIP}}^{(t)}, \bm{V}_{\text{DINOv2}}^{(t)}] \in \mathbb{R}^{d_\text{v}}$ where $d_\text{v} = d_\text{s} + d_\text{d}$.
This concise yet comprehensive feature set effectively captures both high-level semantic and fine-grained visual representations.

\subsection{Aligned Hierarchical Bidirectional Scan module}
Integrating video features into LLMs requires effective mechanisms for capturing spatio-temporal dependencies while managing computational complexity.
The dependencies often involve hierarchical temporal structures, where temporal dynamics occur at multiple timescales.
The AHBS module addresses this challenge by processing visual features through parallel temporal pathways operating
at different sampling rates, allowing the model to capture both fine-grained and coarse temporal dynamics.
This module differs from prior bidirectional scanning methods~\cite{visionmamba,liu2024vmamba} by explicitly modeling the diverse and complex temporal dynamics inherent in videos.

Fig.~\ref{fig:scan} illustrates the architecture of our AHBS module.
The module takes $(\bm{V}^{(1)},\bm{V}^{(2)}, \dots, \bm{V}^{(T)}) \in \mathbb{R}^{T\times N_\text{p}\times d_\text{v}}$ as input.
First, to reduce the computational complexity and focus on temporal modeling, a spatial downsampling layer is applied to the input to obtain $\bm{V}_\text{d} \in \mathbb{R}^{T\times N_\text{d}\times d_\text{v}}$, where $N_\text{d}$ denotes the dimension of the downsampled tokens.
Then, $\bm{V}_\text{d}$ is processed through $M$ parallel pathways, each at a different temporal resolution.
For each pathway, $\bm{V}_m \in \mathbb{R}^{T_\text{m}\times N_\text{d}\times d_\text{v}}$ is obtained by applying temporal downsampling to $\bm{V}_\text{d}$, where $T_m = \lfloor{\frac{T}{2^{m-1}}}\rfloor$ is the temporal downsampling factor for $m\in \{1, \dots, M\}$, with $T_1$ representing the full resolution.

A bidirectional scan module is then applied to $\bm{V}_m$ to model the temporal dependencies.
Integration is applied across multiple resolutions as follows to obtain the output of the module $\bm{H}_\text{v}\in\mathbb{R}^{T\times N_\text{d}\times d_\text{v}}$:
\begin{align}
\bm{H}_\text{v} = \underset{m}{\text{Aggregate}}(\text{SSM}(\bm{V}_m) + \text{SSM}(f_\text{rev}(\bm{V}_m))),
\end{align}
where $\text{Aggregate}(\cdot)$, $\text{SSM}(\cdot)$, and $f_\text{rev}(\cdot)$ denote an aggregation function(e.g., add, concat, interleave), the Mamba operation~\cite{gu2024mamba}, and a function that reverses the input sequence.

\subsection{Mamba-LLM}
We adopt Mamba~\cite{gu2024mamba} as the core backbone LLM,
specifically selected for its efficiency in processing fused video-language sequences.
Leveraging its core selective scan mechanism, Mamba dynamically controls the model dynamics across the input sequence,enabling superior performance over Transformers on certain language modeling tasks.

As input, our Mamba-LLM takes a unified sequence formed by the concatenation of $\bm{H}_\text{vision}$ and the embedded tokens obtained by tokenizing $\bm{x}_\text{txt}$.
The Mamba backbone consists of a stack of identical basic blocks, each comprising a short convolution, a Mamba block, a residual connection, and a normalization layer.
For the detailed formulations of our language backbone, see supplementary.

The output of Mamba-LLM is the token sequence $\hat{\bm{y}} = (\hat{y}_1, \hat{y}_2, \dots, \hat{y}_K)$, where $K$ denotes the sequence length.
At each step $i$, $\hat{y}_i$ is obtained in an auto-regressive manner as $\hat{y}_i = \arg\max_{\tilde{y} \in \mathcal{V}} p_\theta(\tilde{y} | \bm{x}_\text{vision}, \bm{x}_{\text{txt}}, \hat{\bm{y}}_{<i})$, where $\mathcal{V}$ and 
$\hat{\bm{y}}_{<i}$
denote the set of all possible output tokens and the sequence of previously predicted tokens, respectively.
We used the cross-entropy loss as the loss function.
\section{Experiments}

\begin{table*}[t]
  \centering
  \setlength{\tabcolsep}{2pt}
  \resizebox{\textwidth}{!}{%
  \begin{tabular}{
    >{\centering\arraybackslash}p{15mm}
    >{\raggedright\arraybackslash}p{28mm}
    >{\centering\arraybackslash}p{10mm}
    >{\centering\arraybackslash}p{15mm}
    >{\centering\arraybackslash}p{15mm}
    >{\centering\arraybackslash}p{15mm}
    >{\centering\arraybackslash}p{15mm}
    >{\centering\arraybackslash}p{15mm}
    >{\centering\arraybackslash}p{15mm}
    >{\centering\arraybackslash}p{20mm} 
    }
  \toprule
  \multirow{2}{*}{Dataset} &
  \multirow{2}{*}{Method} & 
  \multirow{2}{*}{Size} & 
  \multirow{2}{*}{BLEU1$\uparrow$} & 
  \multirow{2}{*}{BLEU4$\uparrow$} & 
  \multirow{2}{*}{ROUGE$\uparrow$} & 
  \multirow{2}{*}{CIDEr$\uparrow$} & 
  \multirow{2}{*}{METEOR$\uparrow$} & 
  \multirow{2}{*}{PAC-S$\uparrow$} &
  Throughput$\uparrow$\\ 
  &&&&&&&&& (tokens/s)\\
  \hline 
  \multirow{10}{*}{\shortstack{VATEX}}
  &\multicolumn{9}{>{\columncolor{TitleColor}}l}{Proprietary} \\
    &Gemini‑1.5‑Pro          & --     
      & 22.8  & 4.5  & 19.3  & 14.2 & 8.9   & 39.2 & --\\
  &\multicolumn{9}{>{\columncolor{TitleColor}}l}{Fully open MLLMs} \\
    &Video‑ChatGPT           & 7B    
      & 14.8  & 2.4  & 25.0  & 9.9  & 10.6  & 39.4 & 35.2\\
    &Video‑LLaVA             & 7B   
      & 67.3  & 24.5 & 44.9  & 37.7 & 19.6  & 40.8 & 28.4\\
    &LLaVA‑OneVision         & 7B   
      & 60.6  & 17.6 & 41.5  & 39.2 & 21.2  & 42.0 & 17.2\\
  &\multicolumn{9}{>{\columncolor{TitleColor}}l}{Small MLLMs} \\
    &InternVL2.5             & 2.2B 
      & 58.3  & 16.5 & 35.8  & 33.2 & 17.2  & 41.4 & 29.3\\
    &VideoLLaMA3             & 2B   & 56.8
      & 17.5    & 42.2   & 36.9    & \textbf{24.0}   & \textbf{43.0}    & 29.7 \\ 
  \cmidrule(l{0mm}r{0mm}){2-10} 
  \rowcolor{LightCyan}
  \cellcolor{white} 
    &ABMamba (Ours)              & 3.6B 
      & \textbf{73.4} & \textbf{28.6} & \textbf{47.7} & \textbf{44.4} & 22.2 & 41.8 & \textbf{83.8}\\
      
\hline
\multirow{10}{*}{MSR-VTT}
  &\multicolumn{9}{>{\columncolor{TitleColor}}l}{Proprietary} \\
    &Gemini‑1.5‑Pro          & --   & 51.6  & 12.0  & 36.8  & 19.4 & 20.8  & 40.8 & --\\
  &\multicolumn{9}{>{\columncolor{TitleColor}}l}{Fully open MLLMs} \\
    &Video‑ChatGPT           & 7B & 56.3  & 14.4  & 42.4  & 17.8 & 24.8  & 39.4 & 38.1\\
    &Video‑LLaVA             & 7B & 68.0  & 23.3  & 50.1  & 30.7 & 25.7  & 40.8 & 28.9 \\
    &LLaVA‑OneVision         & 7B & 52.5  & 12.4  & 37.4  & 10.8 & 22.8  & 42.5 & 24.8  \\
  &\multicolumn{9}{>{\columncolor{TitleColor}}l}{Small MLLMs} \\
    &InternVL2.5             & 2.2B  & \textbf{69.7}  & 19.1  & 43.0  & \textbf{32.0} & 22.2  & 41.0 & 31.5\\
    &VideoLLaMA3             & 2B & 59.4  & 15.7  & 41.5  & 17.9 & 25.3  & \textbf{42.9} & 33.5 \\ 
  \cmidrule(l{0mm}r{0mm}){2-10} 
  \rowcolor{LightCyan}
  \cellcolor{white}
    &ABMamba (Ours)              & 3.6B & 68.1 & \textbf{23.6} & \textbf{50.6} & 27.3 & \textbf{27.0} & 40.1 & \textbf{95.4}\\
  \bottomrule
  \end{tabular}}
  \vspace{2mm}
  \caption{Quantitative comparison between ABMamba and baseline methods on the test sets of the VATEX and MSR-VTT benchmarks. The best score for each metric is shown in $\textbf{bold}$. We compared our fully open MLLM ($<$4B parameters) against other fully open MLLMs ($<$7B) and small ($<$4B) but not fully open MLLMs. }
  \label{tab:vatex}
\vspace{-5mm}
\end{table*}

\subsection{Experimental Setup}
\subsubsection{Data Details}
\label{subsection:data_details}
We streamlined the training process by removing the pre-alignment phase commonly employed in LLaVA-style paradigms~\cite{liu2024improved,chu2024mobilevlm,zhao2025cobra}.
This phrase addresses the persistent underfitting issues reported in prior work~\cite{karamcheti2024prismatic}.
Instead, we adopted a simplified approach that directly fine-tunes both the vision-language projector and the full LLM backbone.
The fine-tuning was conducted on the 665K Image--Text Instruction dataset introduced in LLaVA 1.5~\cite{liu2024improved}, which comprises diverse supervision signals from COCO, Visual Genome, GQA, and other datasets.
We further incorporated a dataset from Video-ChatGPT~\cite{maaz24acl} including 100K video-text instruction samples, comprising high-quality video-instruction pairs generated through a combination of human-assisted and semi-automatic annotation.

For evaluation, we used the standard video captioning benchmarks MSR-VTT~\cite{MSR-VTT} and VATEX~\cite{wang19iccv}.
Each video was preprocessed by uniformly sampling $T$ frames and resizing them to $384 \times 384$.
The following provides additional details on the benchmarks employed during evaluation:

\paragraph{VATEX~\cite{wang19iccv}.} A multilingual video captioning benchmark comprising 41,269 short video clips covering 600 human activities. All videos were sourced from the Kinetics-600 dataset~\cite{carreira2018shortnotekinetics600}. Each clip was annotated with 10 English and 10 Chinese captions, resulting in 825,380 high-quality descriptions (412,690 per language) collected from over 2,500 annotators. The dataset also included 206,345 English–Chinese parallel sentence pairs, but only the English captions were used in the experiments. 
The average English caption length was 15.23 words, and the vocabulary size was 58,885.

\paragraph{MSR-VTT~\cite{MSR-VTT}.} A benchmark comprising 10,000 web video clips paired with 200,000 clip-sentence annotations. The videos were collected using 257 representative queries from a commercial video search engine, covering 20 diverse categories (e.g., sports, music, cooking) and totaling 41.2 hours in duration. Each 10--30 second clip was annotated with 20 human-written captions, sourced from 1,317 Amazon Mechanical Turk workers. The corpus contained 1,856,523 words and had a vocabulary size of 29,316.

\begin{table*}[t]
\captionsetup{type=figure}
{\small 
  \raggedright
  \renewcommand{\arraystretch}{1}
  \setlength{\tabcolsep}{4pt} 
  \begin{tabular*}{\textwidth}{@{\extracolsep{\fill}}
      m{0.02\textwidth}
      m{0.12\textwidth}
      >{\raggedright\arraybackslash}m{0.40\textwidth}
      >{\raggedright\arraybackslash}m{0.40\textwidth}
    @{}}
    & & (i) & (ii) \\ 
    \toprule
    (a) & $\bm{x}_\text{vision}$ &
      \includegraphics[trim=40 30 40 30,clip,width=0.39\textwidth,height=0.18\textwidth]{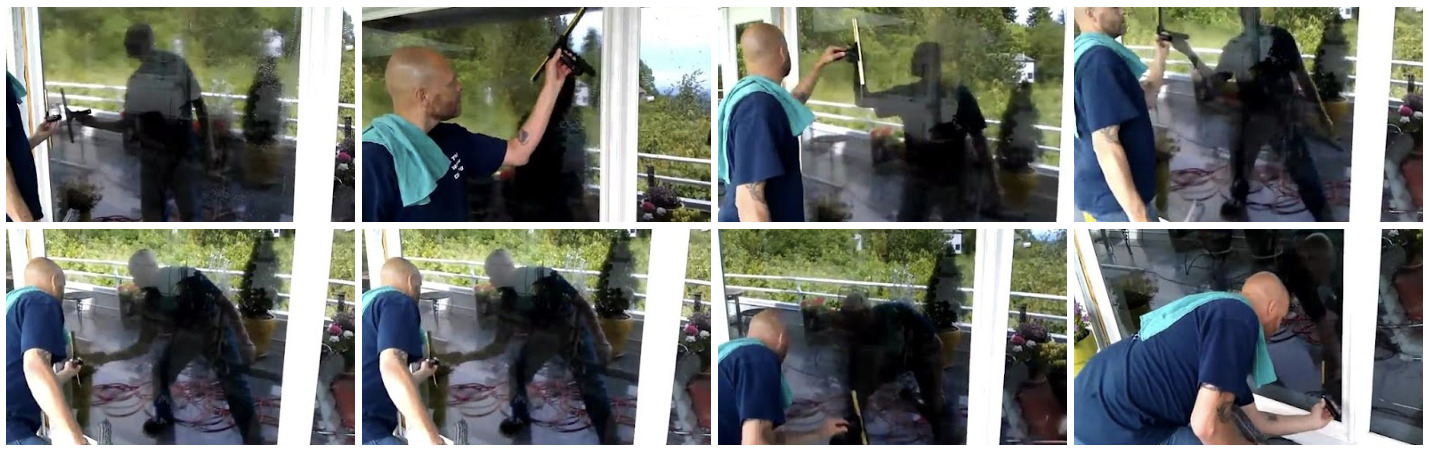} &
      \includegraphics[trim=40 30 40 30,clip,width=0.39\textwidth,height=0.18\textwidth]{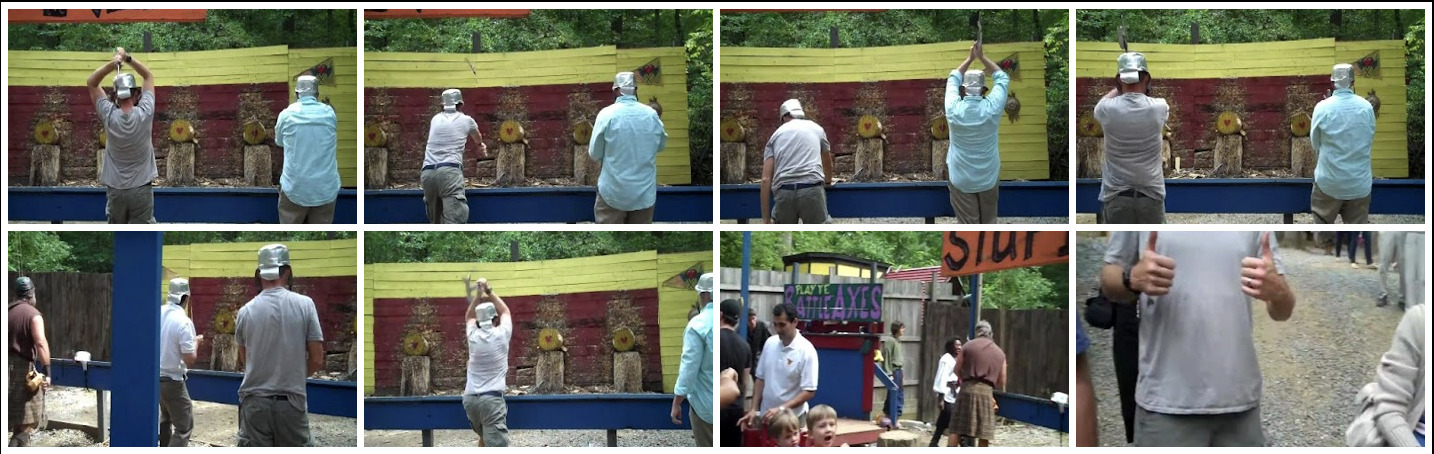} \\
    \midrule
    (b) & $\bm{y}$ &
      A man cleans a window with a squeegee while outside. &
      A group of men are throwing axes at tree stumps several feet away. \\
    \midrule
    (c) & ABMamba &
      A man is cleaning a window with a squeegee. &
      A group of people are standing in a field and throwing axes at a target. \\
    \midrule
    (d) & InternVL2.5 &
      The man sprays the window with a spray bottle. &
      The man wearing a blue shirt is aiming at the board. \\
    \midrule
    (e) & LLaVA-OneVision &
      A man cleans a large glass window, focusing on the lower section. &
      A man gives a thumbs-up sign while standing on a gravel path, surrounded by people in various attire. \\
    \bottomrule
  \end{tabular*}
  \vspace{1mm}
  \caption{Qualitative results of ABMamba and baseline methods on the VATEX benchmark. Rows (a), (b), (c), (d), and (e) show $\bm{x}_\text{vision}$, $\bm{y}$, and the captions generated by ABMamba, InternVL2.5, and LLaVA-OneVision, respectively. Note that $\bm{y}$ denotes one of the reference captions.}
  \label{tab:qual_success}
} 
\vspace{-8mm}
\end{table*}
\subsubsection{Implementation Details}
Mamba-2.8b-zephyr~\footnote{https://huggingface.co/xiuyul/mamba-2.8b-zephyr} was used as the backbone LLM.
The image encoders DINOv2~\cite{oquab24tmlr} and SigLIP ViT-SO~\cite{zhai23iccv} remained frozen during the entire training procedure.
Our model had approximately 3.6B trainable parameters and a total of 2.1T multiply-add operations.
All models were trained using 16 NVIDIA H200 SXM GPUs (VRAM 141GB).
Evaluation of all models was conducted with a single NVIDIA A100 (VRAM 40GB).
The total training time was approximately 8 hours.
See supplementary for further experimental settings.

\subsubsection{Baselines}
We compared ABMamba against several fully open video MLLMs: Video-ChatGPT~\cite{maaz24acl}, Video-LLaVA~\cite{videollava}, and LLaVA-OneVision~\cite{llavaonevision} as well as two small video MLLMs ($<$4B params and not fully open): InternVL2.5~\cite{chen2024internvl} and VideoLLaMA3~\cite{damonlpsg2025videollama3}. 
To prevent out-of-memory errors, we limited the number of frames to 8 for VideoLLaMA3.
Video-ChatGPT, Video-LLaVA, and LLaVA-OneVision were selected as baselines, because they represent fully open MLLMs that have demonstrated strong performance in video understanding tasks.
InternVL2.5 and VideoLLaMA3 were included as they are representative small video MLLMs, with similar numbers of parameters ($<$4B param) to ABMamba.
{While larger models exist, we focus on the 7B scale to conduct a fair and controlled comparison.}

\subsubsection{Evaluation Metrics}
We used standard evaluation metrics for video captioning, including BLEU~\cite{bleu}, ROUGE-L~\cite{rouge}, METEOR~\cite{meteor}, CIDEr~\cite{cider}, and PAC-S~\cite{sarto2023positive}.
We did not use G-VEval because this metric was specifically designed for short-form videos (of less than 10 seconds~\cite{tong2025g}), whereas the video durations in our evaluation benchmarks typically exceeded this threshold (see Section~\ref{subsection:data_details}).

\subsection{Quantitative Results}
\label{sec:caption-level-exp}

Table~\ref{tab:vatex} presents a quantitative comparison between ABMamba and the baseline methods on the VATEX and MSR-VTT benchmarks.
For both benchmarks, we conducted evaluations on their test sets reported based on a single run.
The best score for each metric is highlighted in bold.
We have also included results from Gemini-1.5-Pro as a reference point for proprietary models.

Table~\ref{tab:vatex} indicates that ABMamba was competitive with the baseline methods across most evaluation metrics.
Notably, ABMamba achieved the highest BLEU4 scores of 28.6 on the VATEX benchmark and 23.6 on the MSR-VTT benchmark, outperforming the second-best model by 4.1 and 0.3 points, respectively.

Table~\ref{tab:vatex} also compares the throughput of the baselines and ABMamba.
For this, we randomly selected 10 videos from the VATEX and MSR-VTT benchmarks.
For each benchmark, we averaged the generation speed in tokens per second over the videos.
We set the maximum output length to 512 tokens and measured the total time required from the initiation of video sampling to the completion of caption generation.

From Table~\ref{tab:vatex}, we find that ABMamba achieved an average decoding speed of 95.4 tokens per second on the MSR-VTT benchmark.
Conversely, the fastest baseline method, Video-ChatGPT, which achieved a throughput of 38.1 tokens per second on the same evaluation setup.
This indicates that our approach achieved approximately three times the inference speed of the baseline.
Notably, ABMamba also outperformed InternVL2.5 in terms of decoding speed, despite having more parameters. 
This efficiency gain highlights the effectiveness of our Mamba based approach in handling long video sequences, where the linear complexity of SSMs allows for faster processing than the quadratic complexity of the attention mechanisms typically found in Transformer based models.

\subsection{Qualitative Results}
Fig.~\ref{tab:qual_success} presents qualitative results of ABMamba and two of the baseline methods, InternVL2.5 and LLaVA-OneVision.
In Fig.~\ref{tab:qual_success}, rows (a), (b), (c), (d), and (e) show $\bm{x}_\text{vision}$, $\bm{y}$, and the corresponding video captions generated by ABMamba, InternVL2.5, and LLaVA-OneVision, respectively.
The caption in Fig.~\ref{tab:qual_success} (i)-(c) correctly identified the action and the item as ``A man is cleaning a window with a squeegee,'' demonstrating the ability of ABMamba to capture the primary action occurring in the video.
In constast, Fig.~\ref{tab:qual_success} (i)-(d) incorrectly stated that ``The man sprays the window with a spray bottle'' because Fig.~\ref{tab:qual_success} (i)-(a) shows the man employing a squeegee to wipe the window surface.
Furthermore, the caption in Fig.~\ref{tab:qual_success} (i)-(e) hallucinated by saying ``A man cleans a large glass window, focusing on the lower section,'' while the video frames show the man cleaning the entire window. 

Fig.~\ref{tab:qual_success} (ii)-(c) shows another successful example where ABMamba generated the caption ``A group of people are standing in a field and throwing axes at a target.''
This caption offers a comprehensive representation of the video frames in Fig.~\ref{tab:qual_success} (ii)-(a).
ABMamba correctly captioned the presence of multiple actors (``A group of people''), their primary action (``throwing axes''), and the object of their interaction (``at a target'').
On the other hand, Fig.~\ref{tab:qual_success} (ii)-(d) illustrates that the baseline method inappropriately generated the caption ``The man wearing a blue shirt is aiming at the board.''
This caption, while identifying one actor and their action of aiming towards a ``board'' (presumably the target), failed to describe the primary action (``throwing axes'') of the event depicted in the video frames of Fig.~\ref{tab:qual_success} (ii)-(a).
The caption in Fig.~\ref{tab:qual_success} (ii)-(e) was ``A man gives a thumbs-up sign while standing on a gravel path, surrounded by people in various attire.''
This caption demonstrated a temporal misalignment by focusing on a potentially isolated frame towards the end of the sequence, thus failing to capture the primary activity of the video, i.e., a group of people throwing axes.

\subsection{Ablation Studies}

\paragraph{Scanning Ablation.}
Table~\ref{tab:ablation_scan} presents the impact of comprehensive token scanning methods in the AHBS module. 
We evaluated the following variants: (a) without scan, (b) without backward scan, (c) without downsampling, and (d) ABMamba (full).
Models (a), (b), and (c) consistently underperformed against Model (d) in terms of both the BLEU4 and CIDEr scores across the MSR-VTT and VATEX benchmarks.
In particular, on the MSR-VTT benchmark, Models (a), (b), and (c) performed worse than Model (d) by 16.6, 11.2, and 7.7 CIDEr points, respectively.
Among them, Model (a) exhibits the most substantial performance drop, underscoring the importance of bidirectional scanning in capturing complex spatial dependencies in videos. 
Furthermore, the superior performance of Model (d), which incorporates a multi-resolution structure, suggests that effectively modeling both fine- and coarse-grained temporal patterns is crucial in addressing the heterogeneous and hierarchical nature of the temporal dynamics inherent in videos.


\begin{table}[t]
\centering
\setlength{\tabcolsep}{3pt}

\begin{minipage}{0.55\columnwidth}
\centering
\resizebox{\linewidth}{!}{%
\begin{tabular}{lcccc}
    \toprule
    Model & \multicolumn{2}{c}{MSR-VTT} & \multicolumn{2}{c}{VATEX} \\
    \cmidrule(lr){2-3} \cmidrule(lr){4-5}
    & B4$\uparrow$ & C$\uparrow$ & B4$\uparrow$ & C$\uparrow$ \\
    \midrule
    (a) w/o scan & 13.9 & 10.8 & 17.9 & 24.6 \\
    (b) w/o backward scan & 16.5 & 16.2 & 21.1 & 33.4 \\
    (c) w/o downsampling & 17.9 & 19.7 & 21.9 & 41.7 \\
    (d) ABMamba (full) & \textbf{23.6} & \textbf{27.4} & \textbf{28.6} & \textbf{44.4} \\
    \bottomrule
\end{tabular}}
 \vspace{2mm}
 \caption{Ablation study on token scanning method in the hierarchical bidirectional scan module. B4 and C denote BLEU4 and CIDEr scores, respectively.}
  \label{tab:ablation_scan}
\end{minipage}
\hfill
\begin{minipage}{0.43\columnwidth}
\centering
\resizebox{\linewidth}{!}{%
\begin{tabular}{lccccc}
    \toprule
    Model & \(M\) & Stride & BLEU4$\uparrow$ & CIDEr$\uparrow$ \\
    \midrule
    (i) & 1 & 2 & 21.9 & 41.7 \\
    (ii) & 2 & 2 & 24.0 & 42.4 \\
    (iii) & 3 & 2 & \textbf{28.6} & \textbf{44.4} \\
    (iv) & 3 & 4 & 24.7 & 38.8 \\
    \bottomrule
\end{tabular}}
\vspace{2mm}
\caption{Ablation studies on the number of temporal branches \(M\) and stride in the hierarchical bidirectional scan module.}
\label{tab:ahbs-ablation}
\end{minipage}
\vspace{-6mm}

\end{table}

{
\paragraph{Stride and Downsampling Ablation.}
To evaluate the contribution of multi-resolution temporal modeling in the AHBS module, we conduct ablation studies by varying the number of temporal branches \(M\) and the temporal stride used for downsampling. Experiments were performed on the VATEX benchmark, and results are summarized in Table~\ref{tab:ahbs-ablation}.


As shown in Table~\ref{tab:ahbs-ablation}, increasing the number of temporal branches \(M\) from 1 to 3 consistently improves performance across both BLEU4 and CIDEr metrics. The best performance is achieved with model (iii) where $M$ and stride were $3$ and $2$, respectively. This confirms the effectiveness of capturing multi-resolution temporal dynamics. 
In contrast, increasing the stride to 4 leads to a noticeable drop in performance, indicating that excessive temporal downsampling may discard critical motion information.
These findings provide empirical support for our design choices in the AHBS module. 
}

\section{Conclusion}
In this study, we focused on video captioning tasks by fully open MLLMs, where `fully open' refers to the availability of datasets, code, and trained model weights as open source.
The contributions of this study were as follows. We proposed ABMamba, a fully open Deep SSM-based MLLM designed for efficient temporal modeling of videos, achieving sub-quadratic computational scaling with respect to sequence length. We also introduced a novel AHBS module that processes videos across multiple temporal resolutions, effectively capturing complex and fine-grained temporal dynamics. Despite its efficiency, the proposed method maintains competitive performance across most evaluation metrics and achieves approximately three times faster inference speed compared to baseline methods.


\section{Limitations}
{While this work primarily focuses on video captioning, we have also included evaluation on a representative VideoQA benchmark to broaden the scope of analysis (see supplementary).
However, our exploration of video understanding remains limited to a few core tasks, and further generalization to diverse and open-ended video reasoning scenarios (e.g., temporal grounding, multi-turn dialogue, instructional understanding) remains an important direction for future work.
}

\subsubsection*{Acknowledgements.}
This work was partially supported by JSPS KAKENHI Grant Number 23K28168, JST Moonshot.

\bibliographystyle{splncs04}
\bibliography{icpr26}

\newpage

\section{Deep State Space Models}
\label{appendix:ssm}
Recent advances in Deep SSMs~\cite{gu22iclr,gu2024mamba,mamba2} have demonstrated their remarkable advantages over predominant architectures, including Transformers, across various sequence modeling tasks.
Deep SSMs are inspired by traditional SSMs in continuous systems~\cite{kalman1960new}, which map a one-dimensional function or sequence $\mathbf{x}(t) \in \mathbb{R} \mapsto \mathbf{y}(t) \in \mathbb{R}$
through an $Q$-dimensional hidden state $\mathbf{h}(t)\in\mathbb{R}^Q$as follows:
\begin{align}
   \frac{d\mathbf{h}(t)}{dt} = \mathbf{A}\mathbf{h}(t) + \mathbf{B}\mathbf{x}(t),\label{eq:state} \\ 
   y(t) = \mathbf{C}\mathbf{h}(t) + {D}\mathbf{x}(t),\label{eq:output}
\end{align}
where $\mathbf{A}\in\mathbb{R}^{Q\times Q}$ is the state matrix and $\mathbf{B}\in\mathbb{R}^{Q\times 1}$, $\mathbf{C}\in\mathbb{R}^{Q\times Q}$, and $\mathbf{D}\in\mathbb{R}^{1\times Q}$ are the projection matrices.
Equations~\eqref{eq:state} and ~\eqref{eq:output} are discretized by introducing a timescale parameter $\Delta\in\mathbb{R}_+$  and using the zero-order hold~\cite{zhang07}, resulting in:
 \begin{align}
   \mathbf{h}_k = \mathbf{\bar{A}}\mathbf{h}_{k-1} + \mathbf{\bar{B}}\mathbf{x}_k,\label{eq:state_zoh} \\ 
   y_k = \mathbf{C}\mathbf{h}_k + {D}\mathbf{x}_k,\label{eq:output_zoh}
\end{align}
where $\mathbf{\bar{A}} = \exp(\Delta\mathbf{A})$, $ \mathbf{\bar{B}} = (\Delta \mathbf{A})^{-1}(\exp(\Delta\mathbf{A}) - \mathbf{I})\cdot\Delta\mathbf{B}$.
Similar to RNNs, the recursive time evolution of the internal state described by equations~\eqref{eq:state_zoh} and ~\eqref{eq:output_zoh} hinders direct parallel computation.
To mitigate this, S4~\cite{gu22iclr} reformulates the discrete system defined by equations~\eqref{eq:state_zoh} and ~\eqref{eq:output_zoh} into a convolutional formulation:
\begin{align}
\bar{\mathbf{K}} &= \left( \mathbf{C}\bar{\mathbf{B}} + D,\ \mathbf{C}\bar{\mathbf{A}}\bar{\mathbf{B}} + D, \right. \notag\\
&\quad \left. \ldots,\ \mathbf{C}\bar{\mathbf{A}}^{L-1}\bar{\mathbf{B}} + D \right), \label{eq:kernel}\\
\mathbf{y} &= \bar{\mathbf{K}} * \mathbf{x} \label{eq:conv}
\end{align}
where $L$ denotes the length of the sequence, and
$\mathbf{x} = [x_1, x_2, \ldots, x_k, \ldots] \in \mathbb{R}^L,\quad \mathbf{y} = [y_1, y_2, \ldots, y_k, \ldots] \in \mathbb{R}^L$.
This framework allows Deep SSMs to perform efficient training through the parallelized convolutional formulation (equations~\eqref{eq:kernel}~\eqref{eq:conv}) and enables fast inference via the autoregressive formulation (equations~\eqref{eq:state_zoh}~\eqref{eq:output_zoh}).
Furthermore, Mamba~\cite{gu2024mamba} introduces a selection mechanism in equations~\eqref{eq:state}--\eqref{eq:conv}, whereby the parameters $\mathbf{\bar{A}}, \mathbf{\bar{B}},$ and $\mathbf{\bar{C}}$ are conditioned on the input $\mathbf{x},$ thereby enabling time-varying state transitions.
This design facilitates the evolution of the model dynamics over time, enhancing the expressive capacity of Mamba and leading to performance surpassing that of Transformers on certain language modeling tasks.

{
Motivated by multi-scale time-series architectures~\cite{wu2023timesnet,wang2023timemixer} and recent non-causal adaptations of Deep SSMs to vision tasks~\cite{visionmamba,liu2024vmamba}, we focus on improving temporal reasoning within the projector that bridges the vision encoder and the language model. 
This design choice targets a critical bottleneck in video-language modeling, where effective temporal abstraction must be achieved before alignment with the LLM.
Therefore, the key research contribution is not merely the adoption of Mamba, but the novel architectural design of AHBS, which adapts the capabilities of SSMs to the unique demands of video understanding. 
This distinction is critical for achieving the superior performance and efficiency demonstrated by our method.
}

\section{Related Works}
\paragraph{Benchmarks.}
Recent surveys (e.g.,\cite{nguyen-etal-2024-video}) covering video--language understanding have summarized the development of video captioning benchmarks in terms of domain coverage, annotation style, and temporal granularity.
MSR-VTT~\cite{MSR-VTT} and MSVD~\cite{MSVD} serve as standard open-domain datasets, consisting of short videos and multiple crowd-sourced captions per clip.
VATEX~\cite{wang19iccv} builds upon MSR-VTT by substantially increasing both the diversity of human activities and the number of clip-caption pairs.
In contrast to traditional benchmarks that lack temporal granularity, YouCook2~\cite{YouCook2} and ActivityNet Captions~\cite{ActivityNet} support dense captioning of long-form videos through fine-grained temporal annotation.
Movie-based datasets~\cite{TACoS,M-VAD} provide professionally curated descriptions for short video segments, with large-scale pretraining enabled by instructional video corpora~\cite{HowTo100M}, and fine-grained visual understanding supported by short animated clips~\cite{TGIF}.
These benchmarks collectively support research on both open-ended video description and fine-grained temporal modeling.

\section{Architectural Contributions Beyond Mamba Integration}
We provide additional clarification distinguishing our proposed architecture from a naive combination of Mamba and video understanding.
As shown in Table~2 (a) of the main paper, the variant that directly replaces the temporal modeling component with a simple linear layer fails to perform competitively. 
This setting effectively extends existing Deep SSM based MLLMs~\cite{zhao2025cobra,qiao2024vlmamba}, originally designed for static images, to the video domain without introducing dedicated temporal modeling.
This result highlights that such a naive extension lacks the temporal modeling capacity necessary for video understanding, underscoring the limitations of straightforward adaptations and the necessity for specialized temporal modules such as the AHBS module.

\begin{table}[t]
    \centering
    \begin{tabular}{ll}
        \toprule
        \textbf{Setting} & \textbf{Value} \\
        \midrule
        Batch size       & $128$ \\
        Optimizer        & AdamW \\
        LR schedule      & Cosine decay \\
        Learning rate    & $2\times10^{-5}$ \\
        Epoch            & 2 \\
        Warmup ratio     & 0.03 \\
        Weight decay     & 0.03 \\
        Aggregate & Add \\
        T & 16 \\
        M & 3 \\
        \bottomrule
    \end{tabular}
    \vspace{4mm}
    \caption{Experimental settings of ABMamba.}
    \label{tab:exp_settings}
    \vspace{-4mm}
\end{table}

\section{Additional Implementation Details}
\label{sec:impl_detail}
The experimental setting of ABMamba are listed in Table~\ref{tab:exp_settings}.
We used the following prompt to generate the video captions for evaluation:
``Provide a single-sentence caption that matches the style of the preceding videos.''
Additionally, the following prompt was employed during throughput evaluation: ``Describe the video specifically.''

{\section{Additional Quantitative Results}
\begin{table}[t]
  \centering
  \normalsize
  \setlength{\tabcolsep}{8pt}
  \begin{tabular}{lcc}
    \toprule
    Method & Size & VideoMME (w/o sub)$\uparrow$ \\
    \midrule
    Video-ChatGPT     & 7B    & 28.0 \\
    Video-LLaVA       & 7B    & 30.6 \\
    LLaVA-OneVision   & 7B    & \textbf{40.1} \\
    InternVL2.5       & 2.2B  & 27.6 \\
    VideoLLaMA3       & 2B    & 27.2 \\
    ABMamba (Ours)    & 3.6B  & 29.4 \\
    \bottomrule
  \end{tabular}
  \vspace{4mm}
  \caption{Comparison of VideoMME (without subtitle) scores across various models.}
  \label{tab:videomme}
  \vspace{-4mm}
\end{table}

\subsection{Video QA}
\label{sec:videoqa}
Table~\ref{tab:videomme} shows the qualitative results of ABMamba and the baseline methods on the VideoMME benchmark.
The results were calculated by computing the probability of the full answer choice following the prompt (\textit{cloze format}).
Table~\ref{tab:videomme} shows that ABMamba achieves competitive performance among other baseline models.
While ABMamba only surpasses Video-ChatGPT among the larger models, its performance is comparable to several stronger baselines and notably close to models with significantly larger parameter counts, indicating the effectiveness of our approach given the model size.

\subsection{Memory Efficiency}
\label{appendix:memory}
Table~\ref{tab:appendix-memory} presents a detailed comparison of initial and peak memory usage, memory increase, and token-level throughput on the MSR-VTT benchmark.

\begin{table*}[h]
  \centering
  \resizebox{1.0\textwidth}{!}{%
    \begin{tabular}{lccccc}
      \toprule
      {Method} & {Size} & {Initial Mem.(MB)} & {Peak Mem. (MB)} & {Mem. Increase (MB)} & {Throughput (tokens/s)} \\
      \midrule
      Video-ChatGPT     & 7B   & 13{,}440 & 15{,}505 & 2{,}066  & 38.1 \\
      Video-LLaVA       & 7B   & 28{,}104 & 30{,}652 & 2{,}548  & 28.9 \\
      LLaVA-OneVision   & 7B   & 15{,}813 & 21{,}004 & 5{,}191  & 24.8 \\
      InternVL2.5       & 2.2B &  \textbf{4{,}593} & 12{,}698 & 8{,}105  & 31.5 \\
      Video-LLaMA3      & 2B   &  3{,}751 & 26{,}705 & 22{,}955 & 33.5 \\
      {ABMamba (Ours)} & 3.6B & 7{,}088 & \textbf{7{,}570} & \textbf{482} & \textbf{95.4} \\
      \bottomrule
    \end{tabular}
  }
  \vspace{2mm}
  \caption{Comparison of inference-time memory usage and throughput on MSR-VTT. ABMamba achieves both significantly reduced memory overhead and improved throughput.}
  \label{tab:appendix-memory}
  \vspace{-4mm}
\end{table*}

As shown in Table~\ref{tab:appendix-memory}, ABMamba achieves a substantial reduction in memory overhead, with only 482~MB of additional memory required at inference, representing a 77\% decrease compared to the most memory-efficient transformer-based baseline (Video-ChatGPT, 2{,}066~MB).

These results highlight the architectural efficiency of our method, demonstrating that Deep SSMs can serve as a viable and scalable alternative to transformer-based approaches for video-language understanding. The significantly lower memory footprint, coupled with high decoding efficiency, suggests that ABMamba is well-suited for real-world deployment in latency-sensitive and resource-constrained environments.
}

\section{Additional Qualitative Results}
\begin{table}[t]
\captionsetup{type=figure}
\small
  \raggedright
  \renewcommand{\arraystretch}{1}
  \setlength{\tabcolsep}{4pt} 
  \begin{tabular*}{\linewidth}{@{\extracolsep{\fill}}
      m{0.01\linewidth}
      m{0.20\linewidth}
      >{\raggedright\arraybackslash}m{0.70\linewidth}
    @{}}\\ 
    \toprule
    (a) & $\bm{x}_\text{vision}$ &
      \includegraphics[width=0.98\linewidth,height=0.5\linewidth]{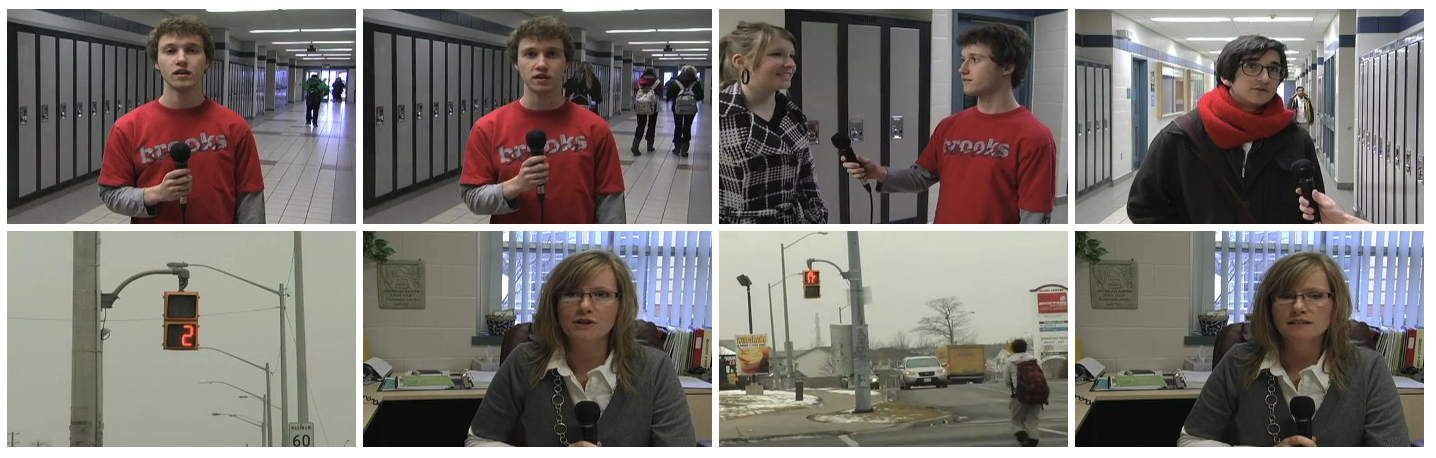} \\
    \midrule
    (b) & $\bm{y}$ &
      A young man is talking on a microphone to the camera, before and after a scene of kids jay-walking in front of cars is shown. \\
    \midrule
    (c) & ABMamba &
      A man in a red shirt talks to a woman in a blue shirt in a hallway. \\
    \midrule
    (d) & InternVL2.5 &
      The man is talking to the camera. \\
    \midrule
    (e) & LLaVA-OneVision &
      A person is seen holding a microphone in a school hallway, with other students walking by. \\
    \bottomrule
  \end{tabular*}
  \vspace{2mm}
  \caption{Failure case from ABMamba and baselines on the VATEX benchmark. Panel (a) shows $\bm{x}_\text{vision}$, while panels (b), (c), and (d) present the captions generated by ABMamba, InternVL2.5, and LLaVA-OneVision, respectively. Note that $\bm{y}$ denotes the reference captions.}
  \label{tab:qual_failure}
  \vspace{-4mm}
\end{table}
Fig.~\ref{tab:qual_failure} illustrates a failure case of ABMamba and two baseline methods on the VATEX benchmark.
In Fig.~\ref{tab:qual_failure}, rows (a), (b), (c), and (d) show $\bm{x}_{\text{vision}}$ and the captions generated by ABMamba, InternVL2.5, and LLaVA-OneVision, respectively.
The video in Fig.~\ref{tab:qual_failure}-(a) features a sequence of multiple interview scenes, in which various individuals (e.g., a man in a red shirt and a woman wearing glasses) are shown holding a microphone and responding to interview questions, interspersed with brief shots of traffic scenes.
However, as shown in Fig.~\ref{tab:qual_failure}, our method generated the caption ``A man in a red shirt talks to a woman in a blue shirt in a hallway,'' omitting both the traffic scene and the segment where the woman with glasses is being interviewed.
Similarly, the baseline methods InternVL2.5 and LLaVA-OneVision, as shown in Fig.~\ref{tab:qual_failure}-(c) and (d) generated limited descriptions: ``The man is talking to the camera'' and ``A person is seen holding a microphone in a school hallway, with other students walking by,'' respectively.
Both captions focus solely on a single interview scene in the hallway.
This failure case indicates that the models predominantly rely on localized visual cues, which may be attributed to the segmented structure of $\bm{x}_{\text{vision}}$, wherein the loosely connected interview and traffic scenes impede the construction of a coherent global narrative.

 \section{Error Analysis}
 \label{sec:error_analysis}
\begin{table}[t]
    \centering
    \begin{tabular}{llc}
        \toprule
        & \textbf{Error category} & \textbf{\#Error} \\
        \midrule
        (i) & Object hallucination       & 68 \\
        (ii) & Action hallucination       & 44 \\
        (iii) & Descriptiveness deficiency & 27 \\
        (iv) & Scene omission            & 24 \\
        (v) & Lexical mismatch           & 9 \\
        \bottomrule
    \end{tabular}
    \vspace{2mm}
    \caption{Categorization of failure modes.}
    \label{tab:failure_modes}
    \vspace{-8mm}
\end{table}

To investigate the limitations of ABMamba, we conducted an error analysis. 
We defined a failure case as a sample for which the CIDEr score was less than that of the sentences generated by a typical baseline (LLaVA-OneVision).
There were 682 and 764 failure cases out of 2,990 and 4,478 samples in the MSR-VTT and VATEX benchmarks, respectively.
Table~\ref{tab:failure_modes} categorizes the failure modes based on the 100 worst failures (of 764 in total) using the VATEX benchmark.
Note that a single failure mode could fall into multiple error categories; therefore, the total count across all error types may exceed 100.

The causes of failure modes can be broadly grouped into five categories:

\paragraph{Object hallucination.} This category refers to modes in which the generated caption mentions an object that does not appear in the input video. A representative example is when the model generated ``A man holding a white bag.'' when the man was actually holding a pillow.
\paragraph{Action hallucination.} This category captures modes in which the generated caption describes an action that is not present in the input video. For example, the model generated ``A person is running.'' although the video only showed the person making gestures while standing still.
\paragraph{Descriptiveness deficiency.} This category refers to modes in which essential objects, actions, or referential expressions are omitted from the caption, resulting in an overly general or under-descriptive caption. A representative example is when the model generated ``The boy is playing with a ball.'' for a scene in which a boy successfully makes a basketball shot and celebrates.
\paragraph{Scene omission.} This category includes modes in which local contexts in the video is treated in isolation, resulting in a loss of overall scene coherence. For instance, in a video depicting a person performing parkour, the model generated captions such as ``A person is jumping'' or ``A person climbing a wall,'' which only capture a localized action without reflecting the broader context.
\paragraph{Lexical mismatch.} This category includes modes in which the generated caption correctly captures the video context but receives an unfairly low evaluation score because of lexical mismatches with the reference captions. A representative example is when the model generated ``A man is cooking at a food stall.'' for a video showing a man cooking in a market, while the reference uses the term ``market'' instead of ``food stall'' despite their semantic similarity. 

Table \ref{tab:failure_modes} indicates that object hallucination errors are the primary bottleneck.
These errors could stem from insufficient integration between visual and language features prior to the language encoder (i.e., late fusion).
A possible solution is to extend the AHBS module to handle both vision and language features within the projector using a mechanism similar to cross-attention.



\end{document}